\documentclass[10pt, a4paper]{article}
\usepackage{lrec2022} 
\usepackage{multibib}
\newcites{languageresource}{Language Resources}
\usepackage{graphicx}
\usepackage{tabularx}
\usepackage{soul}
\usepackage{titlesec}
\titleformat{\section}{\normalfont\large\bfseries\center}{\thesection.}{1em}{}
\titleformat{\subsection}{\normalfont\SmallTitleFont\bfseries\raggedright}{\thesubsection.}{1em}{}
\titleformat{\subsubsection}{\normalfont\normalsize\bfseries\raggedright}{\thesubsubsection.}{1em}{}
\renewcommand\thesection{\arabic{section}}
\renewcommand\thesubsection{\thesection.\arabic{subsection}}
\renewcommand\thesubsubsection{\thesubsection.\arabic{subsubsection}}

\usepackage{epstopdf}
\usepackage[utf8]{inputenc}

\usepackage[whole]{bxcjkjatype} 
\usepackage[unicode]{hyperref} 
\usepackage{xstring}

\usepackage{bm}
\usepackage{multirow}
\usepackage[switch]{lineno}
\usepackage{comment}
\usepackage{longtable}
\usepackage{enumerate}
\usepackage{soul}
\usepackage{xcolor}
\usepackage{gb4e}
 \usepackage{url}
\noautomath

\newcommand{\ctextJA}[3][RGB]{%
  \begingroup
  \definecolor{hlcolor}{#1}{#2}\sethlcolor{hlcolor}%
  \hl{\mbox{#3}}%
  \endgroup
}
\newcommand{\slotJA}[1]{\ctextJA[RGB]{235,235,235}{#1}}

\newcommand{\ctextEN}[3][RGB]{%
  \begingroup
  \definecolor{hlcolor}{#1}{#2}\sethlcolor{hlcolor}%
  \hl{\emph{#3}}%
  \endgroup
}
\newcommand{\slotEN}[1]{\ctextEN[RGB]{235,235,235}{#1}}

\newcommand{\devset}{\textsc{DevSet}}
\newcommand{\evalset}{\textsc{EvalSet}}

\title{TYPIC: A Corpus of Template-Based Diagnostic Comments \\on Argumentation}

\name{Shoichi Naito$^{1,2,5}$, Shintaro Sawada$^{3}$, Chihiro Nakagawa$^{3,2}$,Naoya Inoue$^{4,2,*}$,\\
{\bf \large Kenshi Yamaguchi$^{1}$, Iori Shimizu$^{3}$, Farjana Sultana Mim$^{1}$, Keshav Singh$^{1}$, Kentaro Inui$^{1,2}$}}

\address{
$^{1}$Tohoku University,
$^{2}$RIKEN,
$^{3}$Osaka Prefecture University, 
$^{4}$Stony Brook University,
$^{5}$Ricoh Company, Ltd. \\
         shohichi.naitoh@jp.ricoh.com, \{szb03072, scb03059\}@edu.osakafu-u.ac.jp,\\ chihiro@me.osakafu-u.ac.jp, \{naoya.inoue.lab, keshav.singh29\}@gmail.com\\
         \{kenshi.yamaguchi.e7, inui\}@tohoku.ac.jp, mim.farjana.sultana.t3@dc.tohoku.ac.jp\\}

\abstract{
Providing feedback on the argumentation of the learner is essential for developing critical thinking skills, however, it requires a lot of time and effort.
To mitigate the overload on teachers, we aim to automate a process of providing feedback, especially giving diagnostic comments which point out the weaknesses inherent in the argumentation.
It is recommended to give specific diagnostic comments so that learners can recognize the diagnosis without misinterpretation.
However, it is not obvious how the task of providing specific diagnostic comments should be formulated.
We present a formulation of the task as template selection and slot filling to make an automatic evaluation easier and the behavior of the model more tractable.
The key to the formulation is the possibility of creating a template set that is sufficient for practical use.
In this paper, we define three criteria that a template set should satisfy: expressiveness, informativeness, and uniqueness, and verify the feasibility of creating a template set that satisfies these criteria as a first trial.
We will show that it is feasible through an annotation study that converts diagnostic comments given in a text to a template format.
The corpus used in the annotation study is publicly available.
\\ \newline  }

\begin{document}

\maketitleabstract


\renewcommand*{\thefootnote}{\fnsymbol{footnote}}
\footnotetext[1]{Present affiliation: Japan Advanced Institute of Science and Technology.}
\renewcommand*{\thefootnote}{\arabic{footnote}}

\section{Introduction}
Argumentation and debate are known to be effective tools for developing critical thinking skills~\cite{Roy2005}.
In such argumentation-based education, teachers' feedback helps learners efficiently develop their skills~\cite{Durn2006CriticalTF,Tsui1999COURSESAI,PAULUS1999265}.
Learning becomes more efficient when \emph{specific feedback} is given to the learners~\cite{shute-etal-2008-focus}.
However, it is impractical to ask all teachers to do this because they will need a substantial amount of time to provide specific feedback to each student, and not all teachers have been trained to teach argumentation~\cite{Driver2000}.
It would be highly beneficial to develop a technology that automatically gives specific feedback to students.

\begin{figure}[t]
\centering
\includegraphics[scale=0.119]{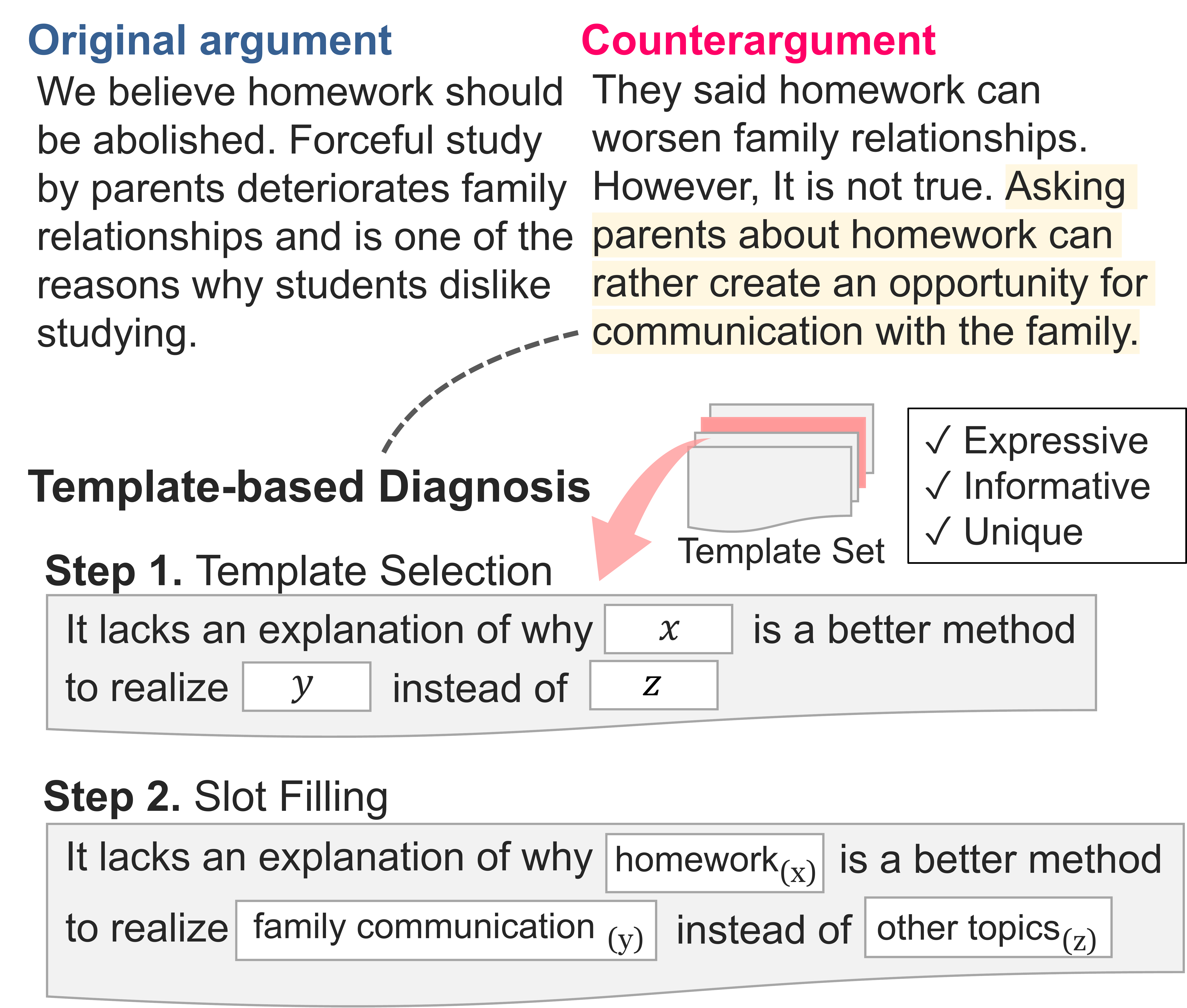}
\caption{Overview of task setting.}
\label{fig:task_overview}
\end{figure}

Essay Scoring aims for assessing students' arguments and giving feedback as a score.
Some studies give a single holistic score for the entire essay~\cite{burstein-chodorow-1999-automated,dong-zhang-2016-automatic}, while others give multi-dimensional scores such as organization, clarity, and justification~\cite{persing-etal-2010-modeling,persing-ng-2013-modeling,persing-ng-2014-modeling,persing-ng-2015-modeling,wachsmuth-etal-2017-computational,carlile-etal-2018-give}.
However, we argue that these score-based feedback is not specific enough for students to develop their skills.
Consider an example argument in Fig.~\ref{fig:task_overview}.
Where we imagine students receive feedback such as \emph{Partially justified: The thesis justifies some of the author's opinions}~\cite{carlile-etal-2018-give}, these students may not know how to revise their arguments because they are not told \emph{how} weak their justification is.

Here, we aim to support learners by automatically giving \emph{more specific diagnostic comments} highlighting the weaknesses of their argumentation.
One challenge with this approach is that it is not clear what task setting should be designed.
One possible approach is to formulate it as a pure generation task, but there is an evaluation issue.
Automatic evaluation metrics such as BLEU~\cite{papineni-etal-2002-bleu} are controversial in generation tasks such as machine translation and dialogue~\cite{liu-etal-2016-evaluate,reiter-2018-structured,mathur-etal-2020-tangled,kocmi2021ship}.
However, it is costly and time-consuming to evaluate generated comments manually.

To address this issue, we propose formulating the task of giving diagnostic comments by \emph{template selection} and \emph{slot filling} as shown in Fig.~\ref{fig:task_overview}.
The task is to select the most likely template from a predefined set of templates and to extract or generate phrases for slots in the selected template.
We assume that diagnostic comments that occur frequently are limited, and having them covered with a predefined set of templates is sufficient for practical use.
Compared with the generation approach, this formulation enables us to use more interpretable evaluation measures such as accuracy, precision, and recall of template selection.
This also helps with error analysis, such as determining which diagnoses the model fails to recognize.

Our research question is the following. Can we create a set of templates that satisfies the following properties?: (i) \emph{expressive}: represent a reasonable amount of common diagnostic comments, (ii) \emph{informative}: preserve the meaning of original comments, and (iii) \emph{unique}: ensure one-to-one mapping between comments and templates.
To investigate this, we collect 1,000 counterarguments and ask people who are experienced in debate education to provide feedback on some of them (\S\ref{sec:data_collection}).
Further, we identify common patterns of feedback to induce a predefined set of templates (\S\ref{sec:template_induction}) and evaluate the quality of the induced set of templates~(\S\ref{sec:annotation_study}).
Our main contributions can be summarized as follows:
\begin{itemize}
\item We propose expressive, informative, and unique templates for argumentative diagnostic comments.
Our study shows that 92.2\% of unseen diagnostic comments can be represented using our templates with moderate inter-annotator agreement (Cohen's Kappa of 0.517), 78\% of which are judged as informative.
\item We publicly release TYPIC corpus\footnote{TYPIC corpus will be publicly available at
\url{https://github.com/cl-tohoku/TYPIC}. The diagnostic comments are given in Japanese, but a translation of them into English will be published. }, consisting of 1,000 counterarguments, 197 of which are annotated with 1,082 diagnostic comments in both natural language and template formats.
\end{itemize}

\begin{figure*}[t]
\centering
\includegraphics[scale=0.48]{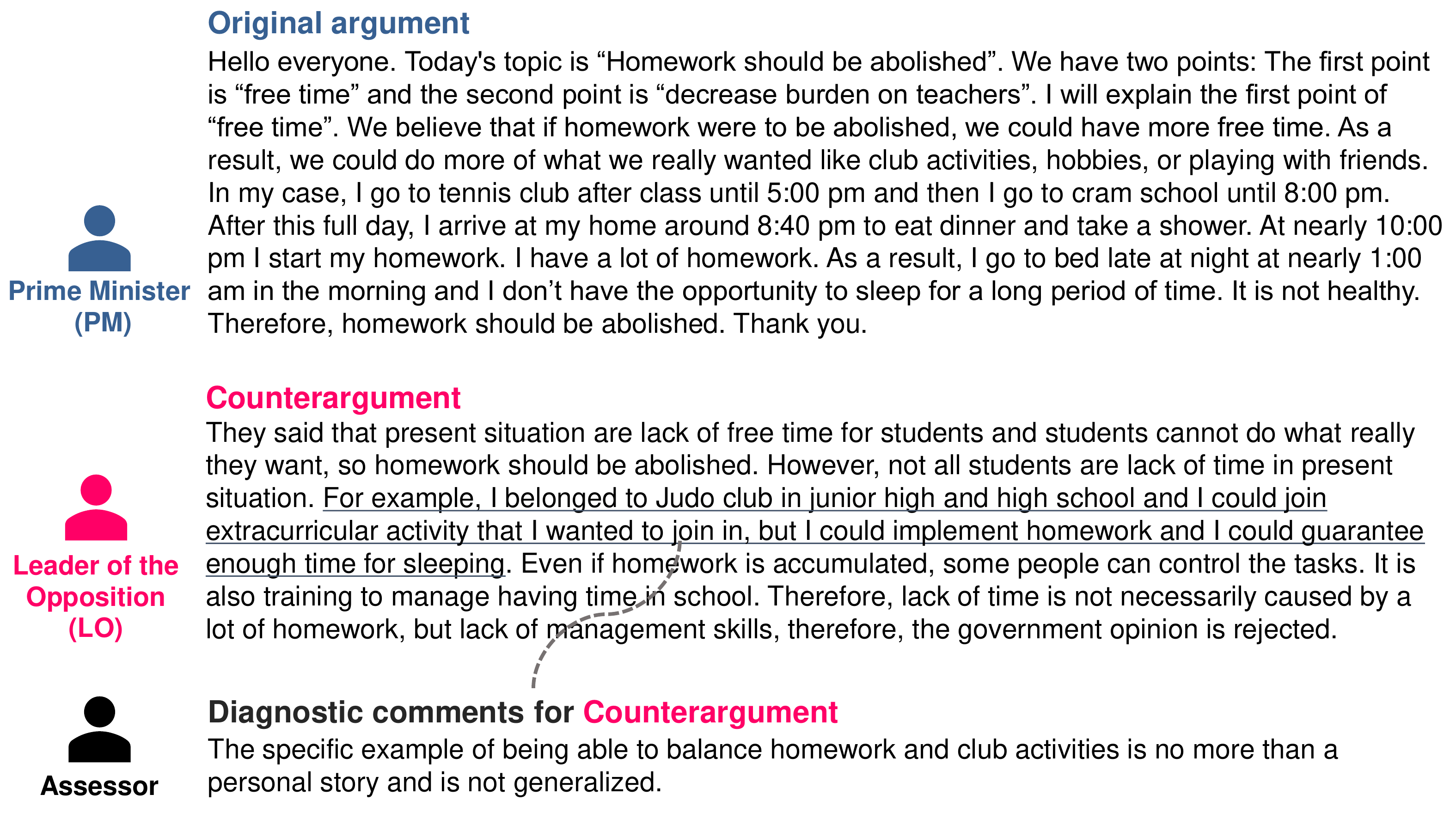}
\caption{An example of parliamentary debate style argumentation and diagnostic comment on counterargument to be collected.}
\label{fig:example_collected_data}
\end{figure*}

\section{Related Work}

\paragraph{Essay scoring}
Essay scoring has been studied as a diagnostic tool for argumentation \cite{Dikli_2006}.
Some studies give a single holistic score for an entire essay ~\cite{burstein-chodorow-1999-automated,dong-zhang-2016-automatic}, while others score for multiple dimensions such as organization~\cite{persing-etal-2010-modeling}, clarity~\cite{persing-ng-2013-modeling}, prompt adherence~\cite{persing-ng-2014-modeling}, and argument strength~\cite{persing-ng-2015-modeling}.
\newcite{wachsmuth-etal-2017-computational} conducted a comprehensive survey on the dimensions of assessing argumentation and created a dataset with scores from 15 dimensions.
Furthermore, \newcite{carlile-etal-2018-give} annotated the scores of multiple dimensions on Persuasive Essay Corpus, and \cite{ke-etal-2019-give} on the ICLE. 
In particular, \newcite{carlile-etal-2018-give} annotated the scores to the fine-grained target of the argumentative discourse unit.
However, these approaches did not consider giving specific diagnostic comments.
Even if a scoring rubric is given as a diagnostic comment, it will produce an abstract comment such as ``Unjustified'' and ``Partially justified.''
Because the reasons for the scores are not entirely clear, it is challenging for learners to understand why and what improvements are required.

\paragraph{Missing premise detection}
Several studies give feedback to indicate the lack of premises for the results of analyzing argumentation structure.
Some studies have proposed an annotation scheme to detect the absence of the premise, considering that a proposition must be supported by an appropriate premise to be evaluable~\cite{park-2015,park-cardie-2018-corpus}.
\newcite{morio2018} and \newcite{egawa-etal-2020-corpus} extended Park's annotation scheme to capture interactions between users for online discussion forums and \newcite{ida-2019-www} proposed an agent system that prompts users to provide additional premises when few premises support a claim. 
\newcite{wambsganss-2020-chi} proposed a support system that analyzes an argumentation made by learners, displays the argumentation structure, and highlights unsupported claims.
However, these approaches cannot provide an in-depth diagnosis of what premise is lacking.

\paragraph{Revision support for argumentative writing}
There exists some research that supports the revision process of argumentative writing.
ArgRewrite automatically classifies the objective of the revision based on the student's draft and revision from eight categories, such as reasoning, rebuttal, evidence, fluency, etc.\cite{zhang-litman-2014-sentence,zhang-litman-2015-annotation,zhang-etal-2016-argrewrite,zhang-etal-2017-corpus,afrin2021}.
By showing the classification results, learners can check whether a revision is consistent with their intentions.
These studies focus on feedback on the revisions made by learners, and not on what revisions should be made.

\paragraph{Recognition of weakness in argumentation}
Some studies recognize the types of weakness in argumentation and give them to learners as feedback.
\newcite{stab-gurevych-2017-recognizing} created a dataset that annotated whether an argument satisfied sufficiency, a criterion evaluated whether premises provided sufficient evidence to accept or reject a claim.
\newcite{ijcai2017-570} defined five error types: grammar error, lack of objectivity, inadequate support, unclear assertion, and unclear justification, as factors that make an argument unpersuasive and annotated the presence of these errors in addition to the holistic score.
However, these approaches can provide the presence of errors, but they do not provide a specific diagnosis that is specific enough for the learner to recognize how to correct them.
The most related work to ours is \newcite{song-etal-2014-applying}．
They proposed a modified Critical Question (CQ) based on Argumentation Scheme \cite{walton_reed_macagno_2008}, which provides specific feedback on the weaknesses inherent in the argumentation.
However, datasets annotated with CQ are not publicly available, and feasibility in terms of expressiveness and semantic change cannot be assessed.
Besides, their annotation scheme of selecting a single segment, such as a sentence or clause, and assigning a static CQ is insufficient to provide specific feedback.
For example, if a CQ is about cause and effect, at least two segments must be selected, cause and effect, otherwise ambiguity remains.
Additionally, diagnosis using phrases not mentioned in the argumentation cannot be expressed.
Our annotation scheme in the form of a template with slots can naturally express both.

\begin{table*}[t]
\centering
\begin{tabular}{lp{13cm}}
\hline
 & \textbf{Homework should be abolished}\\
\hline
HW1 & Abolishing homework gives students more free time \\
HW2 & Forcing students to do homework makes them passive in character \\
HW3 & It is not good for students to be obliged to study by their teachers or parents \\
HW4 & Students have memorized the incorrect way to study with homework \\
HW5 & Schools should take responsibility for the academic skills of children, not parents at home
 \\
\hline
\end{tabular}

\vspace{5mm}
\begin{tabular}{lp{13cm}}
\hline
 & \textbf{Death penalty should be abolished}\\
\hline
DP1 & Death penalty is an inhumane punishment \\
DP2 & Abolishing death penalty will prevent the ending the life of innocent people \\
DP3 & Because of the high stress on the executioner, death penalty should be abolished \\
DP4 & Death penalty deprives criminals of the opportunity for rehabilitation \\
DP5 & The society is brutalized by the use of death penalty \\
\hline
\end{tabular}
\caption{Points of the original argument.}
\label{tab:point_of_pm_speech}
\end{table*}

\section{Dataset}
\label{sec:data_collection}

\subsection{Counterarguments}
We collect counterarguments in the form of parliamentary debate as the target for giving diagnostic comments, as seen in Fig.~\ref{fig:example_collected_data}.
A parliamentary debate is an impromptu debate in which two groups, the government and the opposition, argue about a given topic.
The government takes a position in favor of the topic, while the opposition takes a position against the topic and they give a speech in turn.
This study focuses on the first two speakers, the Prime Minister (PM) on the government side and the Leader of the Opposition (LO) on the opposition side (i.e., original arguments and counterarguments, respectively).

We prepared 10 PM speeches as the original argument on the topics ``Homework should be abolished'' and ``Death penalty should be abolished,'' as shown in Table~\ref{tab:point_of_pm_speech}.
For each of the 10 PM speeches, 100 LO speeches are collected.
Out of 1,000 LO speeches, 250 speeches are written by experienced debaters affiliated with the Parliamentary Debate Personnel Development Association (PDA\footnote{\url{https://pdpda.org/}}).
The remaining 750 speeches are written by Amazon Mechanical Turk workers with Master Qualification.
We pay \$1.60 as a reward per speech.

\begin{table}
\centering
\begin{tabular}{lr}
\hline
\# Counterargument & 1,000 \\
\quad Avg. tokens per argument & 124.0 \\
\quad Avg. sentences per argument & 7.1 \\
\# Diagnostic Comments & 1,082 \\
\quad Avg. \# comments per argument & 5.5\\
\hline
\end{tabular}
\caption{Statistics of TYPIC Corpus.}
\end{table}

\subsection{Diagnostic Comments}

We randomly selected 200 LO speeches (100 for each topic) from the collected counterarguments and asked four assessors to give diagnostic comments on these speeches.
The assessors have at least 4 years of debating and judging experience in high school debate competitions held by PDA.
The assessors read an original argument and counterargument, select the target sentences to be diagnosed, and give a diagnostic comment in natural language sentences.
The instruction is designed to give diagnostic comments on content rather than grammatical errors or expressions to focus on developing thinking skills.
In particular, diagnostic comments are given in terms of relevance to the original argument, justification, and appropriateness of the examples.

All counterarguments are evaluated by two assessors each.
The annotation is conducted under the descriptive paradigm \cite{rottger2021contrasting}, and no instructions are given to ensure that the diagnostic comments of the two assessors agree.
The purpose of adopting this approach is to collect diverse diagnostic comments and analyze the characteristics of this task.

Finally, we divide the collected diagnostic comments into two separate sets in a ratio of 0.25:0.75 for \devset{} and \evalset{}.
We use \devset{} to induce a predefined set of templates, and \evalset{} to evaluate the quality of the created template set.

\begin{table*}
\centering
\small
\begin{tabular}{lp{1.2cm}p{10.9cm}}
\hline
\textbf{Quality Dimension} & \textbf{Category} & \textbf{Template} \\
\hline \noalign{\vskip 1mm}
Local Acceptability & CA2 & なぜ \slotEN{ x } によって \slotEN{ y } という悪い結果が起こるのかが不明確\vspace{-0.5mm}
\newline It is unclear why \slotEN{ x } causes a bad result of \slotEN{ y } \vspace{1mm} \\  
& VAL1 & なぜ \slotEN{ y } にとって \slotEN{ x } が良いことなのかが不明確 \vspace{-0.5mm}
\newline It is unclear why \slotEN{ x } is good for \slotEN{ y } \vspace{1mm} \\
& CLS1 & なぜ \slotEN{ x } は \slotEN{ y } という特性を持つと言えるのかが不明確 \vspace{-0.5mm}
\newline It is unclear why \slotEN{ x } has the property of \slotEN{ y } \vspace{1mm} \\
\hline \noalign{\vskip 1mm}
 & CLS2 & なぜ \slotEN{ z } という点において \slotEN { x } と \slotEN{ y } は同じ/似ているのかが不明確 \vspace{-0.5mm}
\newline It is unclear why \slotEN{ x } and \slotEN{ y } are same/similar in terms of \slotEN{ z } \vspace{1mm} \\ 
Local Sufficiency & EX3 & \slotEN{ x } というのは限定的な状況である \vspace{-0.5mm}
 \newline It is a limiting situation that \slotEN{ x } \vspace{1mm} \\
 & CMP2 &  \slotEN{ y } を実現するのに，なぜ \slotEN{ z } ではなく\slotEN{ x }  という方法が良いのかの説明が不足している \vspace{-0.5mm}
 \newline It lacks an explanation of why \slotEN{ x } is a better method to realize \slotEN{ y } instead of \slotEN{ z } \vspace{1mm} \\
 \hline \noalign{\vskip 1mm}
 Local Relevance & LR1 & なぜ \slotEN{ x } という理由が \slotEN{ y } という結論を支持するのかが不明確 \vspace{-0.5mm}
 \newline It is unclear why a premise \slotEN{ x } supports a claim \slotEN{ y } \vspace{1mm}\\
 \hline \noalign{\vskip 1mm}
Global Relevance & GR2 & 肯定側の \slotEN{ x } という主張に \slotEN{ y } というのは直接的な反論になっていない \vspace{-0.5mm} 
\newline It is not a direct objection to the government's claim \slotEN{ x } to say \slotEN{ y } \vspace{1mm} \\
 \hline \noalign{\vskip 1mm}
 Global Sufficiency & GS1 & なぜ肯定側の \slotEN{ y } という主張よりも否定側の \slotEN{ x } という主張が優位だと言えるのかが不明確 \vspace{-0.5mm} 
 \newline It is unclear why the opposition's claim \slotEN{ x } is superior to the government's claim \slotEN{ y } \vspace{1mm} \\
  & GS2 & 肯定側からの \slotEN{ x } という反論が想定される \vspace{-0.5mm} 
\newline It is expected that the government side will object that \slotEN{ x }  \vspace{1mm} \\
 \hline
\end{tabular}
\caption{An excerpt of the template set used in the annotation study (See Table \ref{tab:template_set_full} in the Appendix for the full version).}
\label{tab:template_set_excerp}
\end{table*}

\section{Inducing Template Set}
\label{sec:template_induction}

As discussed in \S1, we formulate the task of argumentative diagnosis as a template-based task.
We assume that frequently occurring diagnostic comments are limited, and having them covered with a predefined set of templates is sufficient.
Toward this end, we manually induce a predefined set of templates from the collected arguments.

\subsection{Design Choice}
\label{section:criteria}
We assume that an ideal set of templates should satisfy the properties listed below:

\paragraph{Expressive}
It should be able to cover most of the diagnostic comments.
This ensures that learners receive a wide variety of diagnoses.

\paragraph{Informative}
It should preserve the meaning of the original diagnostic comment and maintain the same level of specificity.
This is important for our goal of giving specific diagnostic comments.

\paragraph{Unique}
Only one unique template must be identified for one diagnostic comment.
It is essential to ensure the reliability of the annotations.

The challenge is the difficulty in creating a template set that meets all these criteria.
To see the trade-off between these properties, assume two extreme cases: (i) a template set consisting of only one versatile template (e.g.,  \emph{This argument is unpersuasive.}), and (ii) a template set consisting of very specific templates designed for every single diagnostic comment.
Case (i) satisfies the expressiveness property because the abstract template can represent almost all types of diagnostic comments.
It can also satisfy the uniqueness property since there is only one template.
However, the informativeness is not satisfied because the template-based representation is significantly different from the original diagnosis comment.
In Case (ii), on the other hand, the informativeness is satisfied since the template is identical to the original diagnostic comment.
However, the expressiveness cannot be satisfied because these specific templates are rarely applied to other diagnostic comments.

\subsection{Templates}
\label{sec:templates}
Having the design choice in our mind, we manually designed a template set by analyzing common patterns of the diagnostic comments in \devset{}.
Table \ref{tab:template_set_excerp} summarizes our template set.
Each template consists of a natural language comment and placeholders (henceforth, \emph{slots}).
These slots are intended to be filled with phrases extracted from input arguments or newly generated.
Our templates can be categorized into a standard dimension used in argumentation quality assessment; see \newcite{wachsmuth-etal-2017-computational} and \newcite{wachsmuth-etal-2017-argumentation} for more information.

\subsection{Task Setting}
\label{section:task_definition}

Given the predefined templates, we formulate the task of argumentative diagnosis as two subtasks: \emph{template selection} and \emph{slot filling}, as shown in Fig.~\ref{fig:task_overview}.

\paragraph{Template Selection}
Given (i) an argument, (ii) its counterargument, and (iii) the target argument (indicated by sentences in the counterargument), the task is to identify the target argument's flaw and to choose a template from a list of a predefined set of templates that best reflects the flaw.
We assume that there can be multiple weaknesses for one target argument and consider template selection as a multi-label classification task.
That is, the output is formally defined as a label vector $\bm{l} = (l_1,l_2,\dots,l_n)$, where $l_i \in \{0, 1\}$ indicates whether $i$-th template is an answer or not, and $n$ is the size of the template set.
One can evaluate the models' prediction with evaluation metrics for multi-label classification tasks, such as F1 and accuracy.

\paragraph{Slot Filling}
Given the selected template in the template selection, the task is to fill the slots of the template.
We assume some slot fillers can be extracted from an original argument or counterargument, and some of them must be generated.
One can evaluate the models' prediction with the similarity between the predicted fillers and gold-standard fillers such as n-gram overlaps.
%

\section{Annotation Study}
\label{sec:annotation_study}

We verify the template set described in \S\ref{sec:templates} satisfies the three criteria through an annotation study.
In the annotation study, we annotate the diagnostic comments in \evalset{} with a template using the predefined template set~(\S\ref{sec:annotation_of_applying_template}).
We then use corresponding evaluation metrics to see if the three criteria have been met~(\S \ref{sec:evaluation_metrics}).


\subsection{Annotation Procedure}
\label{sec:annotation_of_applying_template}

We conduct an annotation study to convert diagnostic comments in natural language text into template form using a template set.

The annotation of the template application involves template selection and slot filling.
In the template selection, the annotator selects a template from the template set that expresses the same point as the diagnostic comment.
In slot filling, the annotator fills the slots of the selected template with a phrase.

\begin{exe}
\ex \textbf{Diagnostic comment (natural language):}\\
{\small 宿題を廃止すれば、生徒は性格的に受動的になるという主張の根拠となる理由や例がなく、PMの主張を否定しきれていない}\\
No reasons or examples for the argument that students will become more passive in character if homework is abolished, failing to completely refute the PM's argument.
\end{exe}

For the example above, the diagnosis questions the causality between abolishing homework and the passive personality of students.
In template selection, the appropriate template is CA2, which asks for causality between $x$ and a bad consequence $y$.
The two slots $x$ and $y$ in template CA2 are filled with $x=$\emph{abolishing homework} and $y=$\emph{students becoming passive in character}, respectively.
The final annotation result is as follows.

\begin{exe}
\ex \textbf{Diagnostic comment (templated):}\\
{\small (CA2) \hspace{0.5mm} なぜ \slotJA{ 宿題廃止 } によって \slotJA{ 生徒は性格的} \slotJA{に受動的になる } という悪い結果が起こるのかが不明確}\\
{\small (CA2)} \hspace{1mm} It is unclear why \slotEN{ abolishing homework } causes a bad result of \slotEN{ students becoming passive in character}
\end{exe}

If there is no applicable template in the template set, the annotator selects ``Not Applicable.''

Two annotators who are not authors, annotate the template application.
To calculate inter-annotator agreement (IAA), 74 diagnostic annotations are annotated with overlap between two annotators.
One round of calibration is conducted before the main annotation.

\begin{table}[t]
\centering
\begin{tabular}{cp{6cm}}
\hline
\textbf{Score} & \textbf{Description} \\
\hline
3 & It gives the same diagnosis as the original, without lacking specificity. \\
2 & It gives the same diagnosis, but is less specific. \\
1 & It gives different diagnosis than the original. \\
\hline
\end{tabular}
\caption{Description of the informativeness scores.}
\label{tab:score_semantic_difference}
\end{table}

\subsection{Evaluation Metrics}
\label{sec:evaluation_metrics}
We evaluate how well the template set satisfies the three criteria described in \S\ref{section:criteria}.

\paragraph{Expressiveness} is evaluated using the coverage of the template set for unseen diagnostic comments.
The template set is built based on the observation of 25\% of the diagnostic comments.
We evaluate the coverage of the remaining 75\% of diagnostic comments as a metric.

\paragraph{Informativeness} is evaluated to determine whether it can express the same points as the diagnostic comments in the text without lacking specificity.
We use crowdsourcing to evaluate the template's informativeness manually.\footnote{We used Yahoo! Crowdsourcing (\url{https://crowdsourcing.yahoo.co.jp/}) as a platform.}
We compare the diagnostic comments in the text with those after applying the template and evaluate them using a numerical score from 1 to 3.
Table \ref{tab:score_semantic_difference} shows the rubric.
Each diagnostic comment is judged by five workers, and the results are aggregated by majority voting\footnote{The result of the majority vote is more consistent with the judgment of the experts than that of MACE~\cite{hovy-etal-2013-learning}.
In case of a tie in the majority vote, we count it as a worse score.}.

\paragraph{Uniqueness} is evaluated by IAA of template selection.
This is because if the appropriate template can be uniquely identified, the IAA for template selection should inevitably be high.
We use the agreement rate for the 74 cases where two annotators annotate overlapping as a metric.

\subsection{Results}
\label{sec:evaluation_criteria}

\begin{figure}[t]
\centering
\includegraphics[scale=0.276]{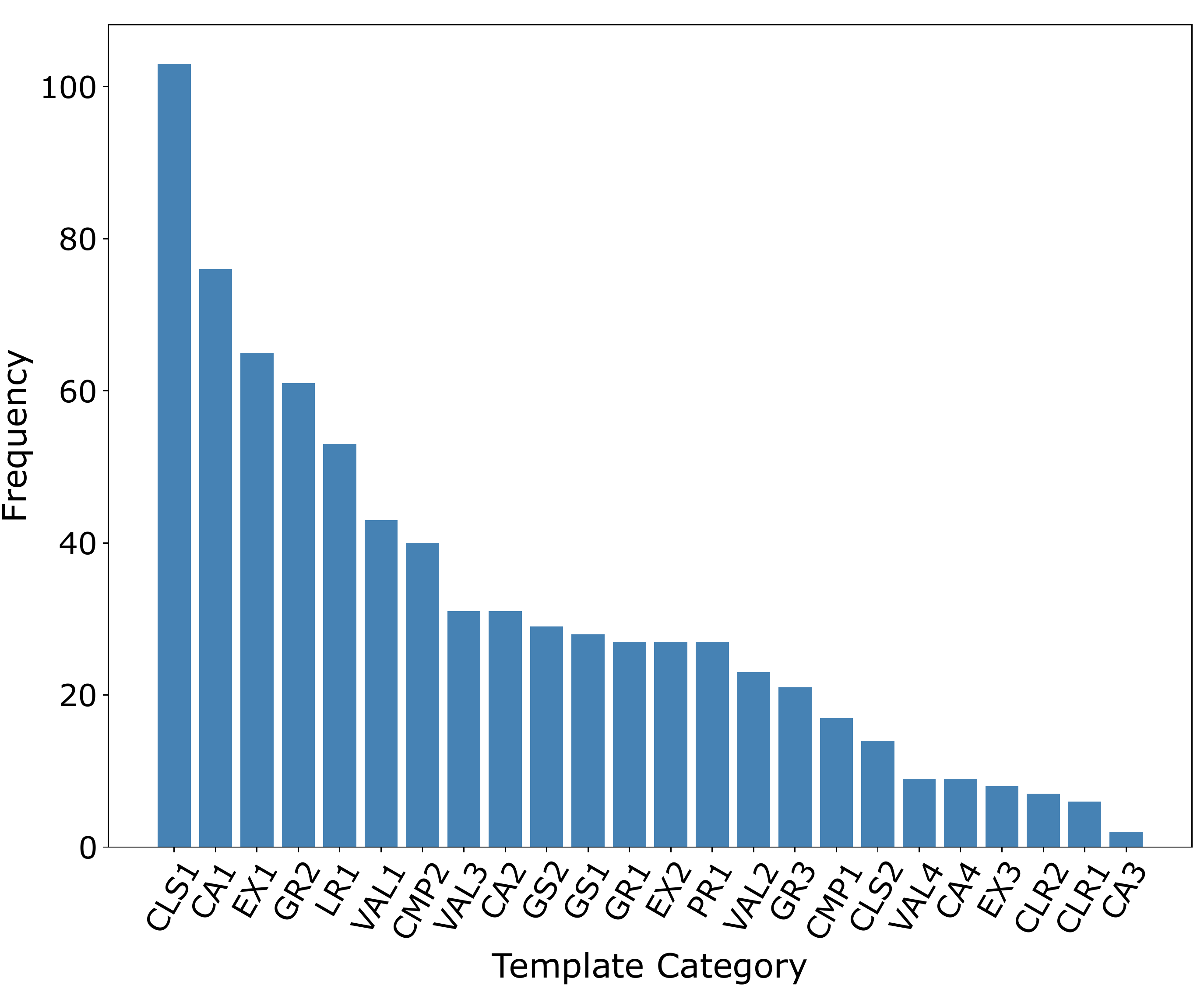}
\caption{Distribution of Template.}
\label{fig:distribution}
\end{figure}

\paragraph{Expressiveness}
The coverage of the template set for unseen comments is $92.2$\% (757/821).
Fig. \ref{fig:distribution} shows the distribution of the templates.
Some examples of diagnostic comments in natural language and those after applying the template are shown in Table \ref{tab:example_annotation_template}.
The result indicates that the types of frequently occurring diagnostic comments are limited and that typical templates can cover numerous diagnostic comments.

\begin{table*}
\centering
\small
\begin{tabular}{lp{6.6cm}p{6.6cm}}
\hline
\textbf{Category} & \textbf{Diagnostic comment in natural language text} & \textbf{Diagnostic comment after applying template}\\
\hline
\multirow{2}{*}{VAL1} & 教育が浅く広い場合の生徒にとってのメリットが分かりにくい
\newline Hard to understand the advantages of shallow and wide education for students.
 & なぜ \slotJA{ 生徒 } にとって \slotJA{ 教育が浅く広いこと } が良いことなのかが不明確
\newline It is unclear why \slotEN{ education being shallow and wide } is good for \slotEN{ students }.\\ 
\hline
\multirow{2}{*}{LR1} & 社会に残虐性は常に存在しているから存在してもいい、は論理が飛躍している
\newline Illogical leap in the point that since brutality has always existed in society, it might as well continue to exist.
& なぜ \slotJA{ 社会に残虐性は常に存在している } という理由が \slotJA{ 存在してもいい } という結論を支持するのかが不明確
\newline It is unclear why a premise \slotEN{``brutalization has always existed in society''} supports a claim \slotEN{``brutalization is acceptable''}.\\ 
\hline
\multirow{2}{*}{CLS1} & なぜ生徒のレベルにあった宿題が出されるのかの理由が述べられていない
\newline No reason given for why homework suited to the students' level is assigned.
& なぜ \slotJA{ 先生が出す宿題 } は \slotJA{ 生徒のレベルに合っ} \slotJA{ている } という特性を持つと言えるのかが不明確
\newline It is unclear why \slotEN{the assignments given by teachers} has the property of \slotEN{matching students' levels}.\\ 
\hline
\multirow{2}{*}{GR2} & PMによる健康や好きなことについての主張に対して成績が落ちるという反論がどう関係するのかが不明瞭である
\newline Unclear how falling grades relate to the PM's arguments about health and leisure.
& 肯定側の \slotJA{ 宿題の廃止により好きなことをして健} \slotJA{康的な生活ができるようになる } という主張に \slotJA{ 宿題の廃止により成績が落ちてしまう } というのは直接的な反論になっていない
\newline It is not a direct objection to the government's claim \slotEN{``without homework, students could do more of what they like and lead healthy lives''} to say \slotEN{``grades will go down if homework is abolished''}.\\ 
\hline
\end{tabular}
\caption{The examples of annotation for template application (See Table \ref{tab:example_annotation_template_appendix} in the Appendix for the others).}
\label{tab:example_annotation_template}
\end{table*}

\begin{figure}[t]
\centering
\includegraphics[scale=0.5]{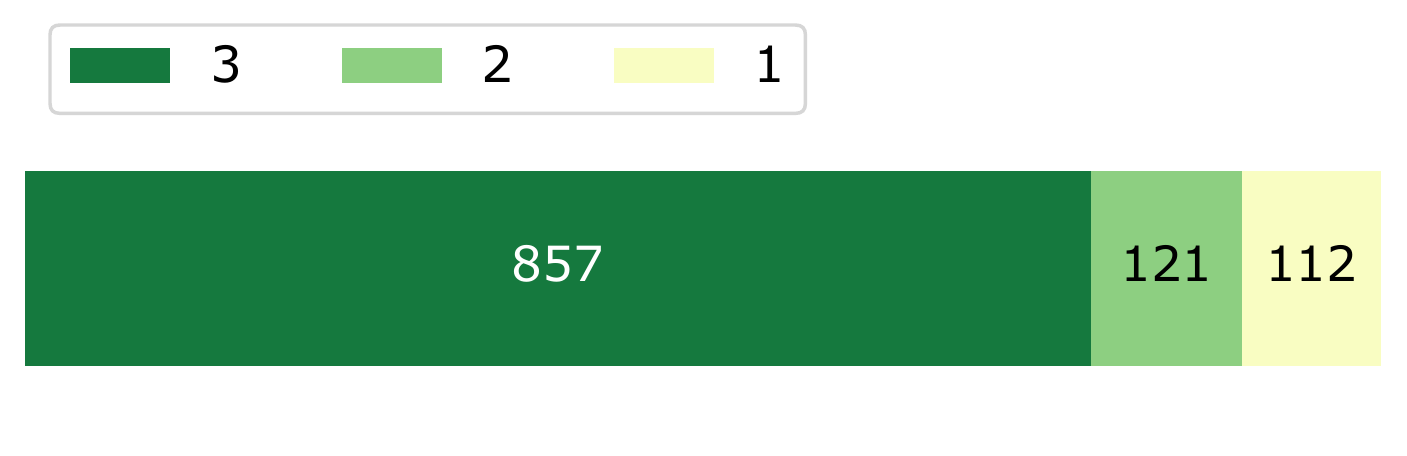}
\vspace{-8mm}
\caption{Distribution of informativeness score.}
\label{fig:specificity}
\end{figure}

\paragraph{Informativeness}
Fig.~\ref{fig:specificity} shows the distribution of the informativeness scores\footnote{The inter-annotator agreement between workers is 0.265 in Krippendorff'$\alpha$ with ordinal distance function \cite{krippendorff80}.}.
The result shows that 78.6\% (857/1090) have the same specificity as a diagnostic comment in the text even after applying a template.
This indicates that template-based diagnostic comments can adequately express the intent of the original diagnosis.

\paragraph{Uniqueness}
The IAA is 0.517 for Cohen's Kappa \cite{Cohen1960}, which corresponds to the moderate agreement \cite{Landis1977}.
Additionally, we evaluate whether the contents of the slots are the same for cases where the template selection is agreed.
The results of the manual evaluation showed that 89\% (65/73) of the slots were substantially consistent in content\footnote{The agreement of the slots is evaluated based on lenient matches. If the fillers of a slot have the same meaning, it is considered as agreed even if the phrases are not exactly the same.}.
These results indicate that annotators can select the appropriate template and fill in the slots with a certain degree of reliability.

Overall, we conclude that it is feasible to create a template set that satisfies three criteria of expressiveness, informativeness, and uniqueness.

\subsection{Analysis}
\subsubsection{Characteristics of Fillers}
\label{sec:charasteristics_fillers}
To analyze the difficulty of slot filling, we randomly select some diagnostic comments and analyze whether or not fillers can be extracted from the original argument or counterargument.
Table \ref{tab:analysis_filler} shows the proportion of fillers that can be extracted from an original argument or counterargument.

For 75.9\% (126/166) of the fillers, they can be extracted from the arguments with minor modifications such as changing the part of speech and paraphrasing.

For 8.4\% of the fillers, the basic phrases can be extracted, but require substantial changes.
The following is a case where the extracted concepts need to be combined.

\begin{exe}
\ex
\textbf{Counterarugment (excerpt):}\\
That is to say even if abolishing homework, students become passive in character. This is because students are instructed by teachers in club activity or cram school in many situation.
\vspace{2mm}
\\
\textbf{Diagnostic comment (templated):}\\
{\small (CLS2)} \hspace{1mm} It is unclear why \slotEN{passivity due to homework} and \slotEN{passivity due to club activities} are same/similar.
\end{exe}

Consider the second filler, ``passivity due to club activity''.
The phrases ``passive in character'' and ``club activity'' can be extracted from the counterargument, but they need to be combined to make the filler.

For 15.7\% of the fillers, it cannot be extracted from an original argument or counterargument. %
Some fillers belong to this case for diagnoses that provide the learner a new perspective.

\begin{exe}
\ex
\textbf{Diagnostic comment (templated):}\\
{\small (CMP2)} \hspace{1mm} It lacks an explanation of why \slotEN{ homework } is a better method to realize \slotEN{students' mastery of basic skills} instead of \slotEN{independent study}.
\end{exe}

In the above case, the third filler, ``independent study,'' which is taken up as a comparison, is not mentioned in the original argument or counterargument.

\begin{exe}
\ex
\label{ex:GS2}
\textbf{Diagnostic comment (templated):}\\
{\small (GS2)} \hspace{1mm} It is expected that the government will object that \slotEN{the death penalty is still one of the factors in the brutalization of modern society}.
\end{exe}

Similarly, the filler of the template for the anticipated objection (GS2) is not extracted, because it highlights aspects that have not been considered in the counterargument (see Example~(\ref{ex:GS2})).

This result suggests that the approach to extracting fillers from the arguments can cover most fillers.
We think this characteristic of the filler will increase the feasibility of automated models.
Despite recent breakthroughs in pre-trained language model, it is still challenging to generate argumentative knowledge with reasoning~\cite{saha-etal-2021-explagraphs}.
The approach to extracting fillers can alleviate the difficulty of the problem.

\subsubsection{Different Diagnoses for the Same Target}
\label{sec:multiple_valid_diagnosis}
To analyze whether the template selection should be a single-label classification or a multi-label classification, we examine how many cases in which the same target sentences have been given different templates.
Table \ref{tab:analysis_count_of_diagnostic_comments_per_target} illustrates the percentage of different templates that were given to the target sentences.

For 28.9\% of the cases, two to five different diagnoses (templates) are given for the same target.
The following is an example of such a case.

\begin{exe}
\ex
\textbf{Counterarugment (excerpt):}\\
Training through homework can cultivate students' perspectives, values, and curiosities.
\vspace{2mm}
\\
\textbf{Diagnostic comment 1 (templated):}\\
{\small (CA1)} \hspace{1mm} It is unclear why \slotEN{homework} causes a good result of \slotEN{ cultivating students' curiosity}.
\vspace{2mm}
\\
\textbf{Diagnostic comment 2 (templated):}\\
{\small (CMP2)} \hspace{1mm} It lacks an explanation of why \slotEN{ homework } is a better method to realize \slotEN{cultivating perspectives, values, and curiosity} instead of \slotEN{other methods}.
\end{exe}

Diagnostic comment 1 questions the belief that homework cultivates students' curiosities.
Alternatively, diagnostic comment 2 questions whether homework is more suitable than other methods to cultivate students' curiosities.
These diagnostic comments differ in how the relationship between homework and students' curiosities is viewed, but both are reasonable.

The results indicate that there are multiple reasonable diagnoses for the same target sentence.
Therefore, we adopt multi-label classification as a task setting of template selection.

\begin{table}[t]
\centering
\begin{tabular}{p{3.8cm}rr}
\hline
 & \multicolumn{2}{c}{Percentage}\\
\hline
Extractable & 75.9\% & ( 126 / 166 ) \\
Extractable {\small (some essential changes are required)} & 8.4\% & ( 14 / 166 ) \\
Not extractable & 15.7\% & ( 26 / 166 ) \\
\hline
\end{tabular}
\caption{Percentage of filler that can be extracted from original argument or counterargument.}
\label{tab:analysis_filler}
\end{table}

\begin{table}[t]
\centering
\begin{tabular}{crr}
\hline
\# Different diagnoses & \multicolumn{2}{c}{Percentage} \\
\hline
1 & 71.1\% & ( 542 / 762 ) \\
2 & 18.9\% & ( 144 / 762 ) \\
3 & 6.8\%  & ( 52 / 762 ) \\
4 & 2.5\%  & ( 19 / 762 ) \\
5 & 0.7\%  & ( 5 / 762 ) \\
\hline
\end{tabular}
\caption{Percentage of different diagnoses given to target sentences.}
\label{tab:analysis_count_of_diagnostic_comments_per_target}
\end{table}

\section{Discussion}

\paragraph{Limitations}
We did not verify whether the template set satisfies the three criteria for diagnostic comments on different topics and given by different assessors.
Whether it can be generalized to other conditions should be explored in the future.
Furthermore, there are still some issues in the current template set.
Template CLS1 is applicable in many cases, and there is room for further subdivisions.
After analyzing the disagreement cases, templates VAL1 and VAL4 are often confused, and they may need to be consolidated into one.
Although some issues still need to be addressed, we believe they can be effectively resolved by repeating the error analysis.

\paragraph{Data Collection of Counterargument}
This study collects counterarguments intensively by focusing on a few topics.
We think it would be difficult to adequately address this task with a broad and shallow approach to collecting counterarguments.
To present appropriate diagnostic comments, it is necessary to analyze the argumentation structure of an original argument and counterargument accurately.
Although pre-trained language models have improved the performance, it is still challenging to analyze argumentation structures with implicit relations \cite{atwell-etal-2021-discourse}.
We aim to alleviate the difficulty by covering the counterarguments that typically appear in a topic.

Of course, the cost of expanding to other topics will be higher than a broad and shallow approach to collecting argumentation.
However, if we think about actual usage scenarios, there will be many situations where only a few topics will be immediately useful.
For example, in a high school classroom, five topics may be sufficient.
It is not necessary to have different topics for each class, nor is it necessary to change them yearly.

\paragraph{Data Collection of Diagnostic Comments}
To collect various diagnostic comments and analyze the task's characteristics, we collected diagnostic comments based on the descriptive paradigm, which does not adjust for agreement among assessors.
It is not suitable for training a model to make consistent predictions.
Based on the results of this trial, we plan to refine the corpus to make it more suitable for model training and evaluation.

\section{Conclusion}
We proposed a template-based formulation to make the task of giving specific diagnostic comments on argumentation more tractable.
As a first attempt, we verified the feasibility of creating a template set that satisfies three criteria of \emph{expressiveness}, \emph{informativeness}, and \emph{uniqueness}.
We showed it is feasible through an annotation study that converts diagnostic comments given in text into a template format.
We publish the corpus used for the annotation study.
The corpus consists of 1,000 counterarguments, 197 of which are annotated with 1,082 diagnostic comments in both natural language and template formats.

Our future work is to refine the corpus to be more suitable for model training and evaluation based on this annotation study.
For modeling, we plan to analyze what kind of information needs to be captured by the model to give a correct diagnostic comment.

\section{Acknowledgments}
This work was partly supported by JSPS KAKENHI Grant Number 22H00524 and NEDO JP1234567.
The authors would like to thank Paul Reisert, members of the
Tohoku NLP Lab, and the anonymous reviewers for their helpful feedback.
We also would like to thank the assessors and the annotators for their time and effort.

\section{Bibliographical References}\label{reference}

\bibliographystyle{lrec2022-bib}
\bibliography{lrec2022-example}


\section*{Appendix: }
Table \ref{tab:template_set_full} shows the template set of the full version used in the annotation study.
Table \ref{tab:example_annotation_template_appendix} shows the examples of annotation for template application.

\begin{table*}
\centering
\begin{tabular}{lp{1.2cm}p{10.9cm}}
\hline
\textbf{Quality Dimension} & \textbf{Category} & \textbf{Template} \\
\hline \noalign{\vskip 1mm}
Local Acceptability & CA1 & なぜ \slotEN{ x } によって \slotEN{ \it{y} } という良い結果が起こるのかが不明確 \vspace{-0.5mm}
\newline It is unclear why \slotEN{ x } causes a good result of \slotEN{ y } \vspace{1mm} \\ 
 & CA2 & なぜ \slotEN{ x } によって \slotEN{ y } という悪い結果が起こるのかが不明確 \vspace{-0.5mm}
\newline It is unclear why \slotEN{ x } causes a bad result of \slotEN{ y } \vspace{1mm} \\  
 & CA3 & なぜ \slotEN{ x } によって \slotEN{ y } という良い結果が抑制されるのかが不明確 \vspace{-0.5mm}
\newline It is unclear why \slotEN{ x } suppresses a good result of \slotEN{ y } \vspace{1mm} \\ 
 & CA4 & なぜ \slotEN{ x } によって \slotEN{ y } という悪い結果が抑制されるのかが不明確 \vspace{-0.5mm}
\newline It is unclear why \slotEN{ x } suppresses a bad result of \slotEN{ y } \vspace{1mm} \\ 
 & VAL1 & なぜ \slotEN{ y } にとって \slotEN{ x } が良いことなのかが不明確 \vspace{-0.5mm}
\newline It is unclear why \slotEN{ x } is good for \slotEN{ y } \vspace{1mm} \\  
 & VAL2 & なぜ \slotEN{ y } にとって \slotEN{ x } が悪いことなのかが不明確 \vspace{-0.5mm}
\newline It is unclear why \slotEN{ x } is bad for \slotEN{ y } \vspace{1mm} \\ 
 & VAL3 & なぜ \slotEN{ x } は \slotEN{ y } とすべきと考えているのかが不明確 \vspace{-0.5mm}
\newline It is unclear why \slotEN{ x } should be \slotEN{ y } \vspace{1mm} \\ 
 & VAL4 & なぜ \slotEN{ x } は \slotEN{ y } とすべきでないと考えているのかが不明確 \vspace{-0.5mm}
\newline It is unclear why \slotEN{ x } should not be \slotEN{ y } \vspace{1mm} \\
 & CLS1 & なぜ \slotEN{ x } は \slotEN{ y } という特性を持つと言えるのかが不明確 \vspace{-0.5mm}
\newline It is unclear why \slotEN{ x } has the property of \slotEN{ y } \vspace{1mm} \\
 & CLS2 & なぜ \slotEN{ z } という点において \slotEN { x } と \slotEN{ y } は同じ/似ているのかが不明確 \vspace{-0.5mm}
\newline It is unclear why \slotEN{ x } and \slotEN{ y } are same/similar in terms of \slotEN{ z } \vspace{1mm} \\ 
 & PR1 & なぜ \slotEN{ x } を実現可能なのかが不明確 \vspace{-0.5mm}
\newline It is unclear why \slotEN{ x } can be feasible \vspace{1mm} \\ 
 \hline \noalign{\vskip 1mm}
Local Sufficiency & EX1 & \slotEN{ x } の例として具体的には何があるか \vspace{-0.5mm}
\newline It lacks the specificity of what exactly is an example of \slotEN{ x } \vspace{1mm} \\ 
 & EX2 & \slotEN{ x } はどの程度 \slotEN{ y } かの具体性に欠ける \vspace{-0.5mm} 
\newline It lacks the specificity regarding the extent to which \slotEN{ x } \slotEN{ y } \vspace{1mm} \\ 
 & EX3 & \slotEN{ x } というのは限定的な状況である \vspace{-0.5mm}
 \newline It is a limiting situation that \slotEN{ x } \vspace{1mm} \\ 
 & CMP1 & なぜ \slotEN{ y } よりも \slotEN{ x } を優先すべきかの説明が不足している \vspace{-0.5mm}
 \newline It lacks an explanation of why \slotEN{ x } should be preferred over \slotEN{ y } \vspace{1mm}\\ 
 & CMP2 &  \slotEN{ y } を実現するのに，なぜ \slotEN{ z } ではなく\slotEN{ x }  という方法が良いのかの説明が不足している \vspace{-0.5mm}
 \newline It lacks an explanation of why \slotEN{ x } is a better method to realize \slotEN{ y } instead of \slotEN{ z } \vspace{1mm} \\
 \hline \noalign{\vskip 1mm}
 Local Relevance & LR1 & なぜ \slotEN{ x } という理由が \slotEN{ y } という結論を支持するのかが不明確 \vspace{-0.5mm}
 \newline It is unclear why a premise \slotEN{ x } supports a claim \slotEN{ y } \vspace{1mm}\\
 \hline \noalign{\vskip 1mm}
 Clarity & CLR1 &  \slotEN{ x } という表現が何を意味しているのか分からない \vspace{-0.5mm}
 \newline It is hard to understand what the statement \slotEN{ x } is means \vspace{1mm} \\ 
  & CLR2 & \slotEN{ x } について具体例はあるが一般化した説明がない \vspace{-0.5mm}
  \newline There is a specific example for \slotEN{ x }, but it lacks a generalized justification \vspace{1mm} \\ 
\hline \noalign{\vskip 1mm}
 Global Relevance & GR1 & \slotEN{ x } という主張/理由が論題とどのように関係するのかが不明確 \vspace{-0.5mm}
\newline It is unclear how the statement \slotEN{ x } relates to the topic \vspace{1mm}  \\ 
  & GR2 & 肯定側の \slotEN{ x } という主張に \slotEN{ y } というのは直接的な反論になっていない \vspace{-0.5mm} 
\newline It is not a direct objection to the government's claim \slotEN{ x } to say \slotEN{ y }  \vspace{1mm} \\ 
  & GR3 & \slotEN{ x } というのは肯定側が定義している \slotEN{ y } を考慮できていない \vspace{-0.5mm} 
\newline The statement \slotEN{ x } fails to consider \slotEN{ y }, which is the definition of the government side  \vspace{1mm} \\ 
  \hline \noalign{\vskip 1mm}
 Global Sufficiency & GS1 & なぜ肯定側の \slotEN{ y } という主張よりも否定側の \slotEN{ x } という主張が優位だと言えるのかが不明確 \vspace{-0.5mm} 
 \newline It is unclear why the opposition's claim \slotEN{ x } is superior to the government's claim of \slotEN{ y } \vspace{1mm} 
 \\ 
  & GS2 & 肯定側からの \slotEN{ x } という反論が想定される \vspace{-0.5mm} 
\newline It is expected that the government will object that \slotEN{ x }  \vspace{1mm} \\
\hline
\end{tabular}
\caption{The template set used in the annotation study (Full version).}
\label{tab:template_set_full}
\end{table*}

\begin{table*}
\centering
\small
\begin{tabular}{lp{6.6cm}p{6.6cm}}
\hline
\textbf{Category} & \textbf{Diagnostic comment in natural language text} & \textbf{Diagnostic comment after applying template}\\
\hline
\multirow{2}{*}{CA4} & 死刑が十分な抑止力になるという根拠が示されていない
\newline No evidence is shown that the death penalty is a sufficient deterrent.
 & なぜ \slotJA{ 死刑 } によって \slotJA{ 犯罪 } という悪い結果が抑制されるのかが不明確 
\newline It is unclear why \slotEN{the death penalty} suppresses a bad result of \slotEN{crime}.\\
\hline
\multirow{2}{*}{VAL3} & なぜ、学校でたくさんのスキルを学べるよう担保するべきなのかが述べられていない
\newline No discussion of why it should be guaranteed that children can learn many kinds of skills at school.
 & なぜ \slotJA{学校} は \slotJA{生徒がたくさんのスキルを学べる} \slotJA{よう担保} すべきと考えているのかが不明確
\newline It is unclear why \slotEN{schools} should be \slotEN{responsible for guaranteeing that students learn many skills}.\\
\hline
\multirow{2}{*}{PR1} & 死刑執行のボランティアは本当に自由意志で募ることができるのかという点が疑問として残る
\newline Doubts remain over whether volunteer executioners would really apply out of their own free will.
& なぜ \slotJA{ 死刑執行のボランティアを本当に自由意} \slotJA{志で募ること }  を実現可能なのかが不明確
\newline It is unclear why \slotEN{recruiting volunteer executioners who actually do it out of their own free will} can be feasible.\\ 
\hline
\multirow{2}{*}{EX1} & 「生徒のスキル」や「生徒の習得状況に合わせた授業の内容」という部分が抽象的
\newline Abstract use of phrases like ``students' skills'' and ``class content suited to students' acquisition of learning.''
& \slotJA{「生徒のスキル」や「生徒の習得状況に合わせ} \slotJA{ た授業」} の具体例には何があるか
\newline It lacks the specificity of what exactly is an example of \slotEN{``students' skills'' and ``class content suited to students' acquisition''}.\\  
\hline
\multirow{2}{*}{EX2} & 犯罪者を刑務所に入れておくにはどれくらいコストがかかるのかという説明がない
\newline No explanation exists of the costs required to keep criminals in prison.
& \slotJA{犯罪者を刑務所に入れておくに} はどの程度 \slotJA{コストがかかる} かの具体性に欠ける
\newline It lacks the specificity regarding the extent to which \slotEN{keeping criminals in jail} \slotEN{costs money}\\  
\hline
\multirow{2}{*}{EX3} & 宿題と部活動の両立ができたという具体例は体験談にとどまっており、一般化されていない
\newline The specific example of being able to balance homework and club activities is no more than a personal story and is not generalized. 
& \slotJA{宿題と部活動を両立できる} というのは限定的な状況である
\newline It is a limiting situation that \slotEN{ being able to balance homework and club practice } \\  
\hline
\multirow{2}{*}{CMP1} &  勉強の好きな生徒のために勉強が嫌いな生徒のことを無視しても良いのかという理由づけがない
\newline No reasoning for why students who do not like studying can be ignored for the sake of those who do.
& なぜ \slotJA{勉強が嫌いな生徒} よりも \slotJA{勉強が好きな} \slotJA{生徒} を優先すべきかの説明が不足している
\newline It lacks an explanation of why \slotEN{students who like studying} should be preferred over \slotEN{students who dislike studying}.\\ 
\hline
\multirow{2}{*}{GR1} &  犯罪を犯した者を無力化する方法としても監獄は有効であるという主張は、論題と関係ない説明に見える
\newline The argument that incarceration is effective to incapacitate those who commit crimes does not seem relevant to the topic.
& \slotJA{犯罪を犯した者を無力化する方法としても監獄} \slotJA{は有効である} という主張/理由が論題とどのように関係するのかが不明確
\newline It is unclear how the statement \slotEN{``incarceration is a way to incapacitate those who commit crimes''} relates to the topic.\\
\hline
\multirow{2}{*}{GS1} &  リスクがゼロでないことを問題として挙げているのに対し、なぜ最小限にするので十分なのかの説明が不足している
\newline Insufficient explanation exists regarding why minimizing risk is sufficient in response to the point that the risk exists.
& なぜ肯定側の \slotJA{ リスクがゼロでないことが問題 } という主張よりも否定側の \slotJA{ リスクを最小限} \slotJA{にするので十分} という主張が優位だと言えるのかが不明確
\newline It is unclear why the opposition's claim \slotEN{``minimizing the risk is enough''} is superior to the government's claim of \slotEN{``the problem raised is that the risk exists''}.\\
\hline
\multirow{2}{*}{GS2} &  授業中に宿題をするということは人によっては悪いことだと思ってしまう可能性がある
\newline Some people might consider it a bad thing to do homework during class.
& 肯定側から \slotJA{ 授業中に宿題をするのは悪いことだ } という反論が想定される
\newline It is expected that the government will object that \slotEN{doing homework during class is bad}.\\ 
\hline
\end{tabular}
\caption{The examples of annotation for template application.}
\label{tab:example_annotation_template_appendix}
\end{table*}

\end{document}